\documentclass[10pt,twocolumn,letterpaper]{article}

\usepackage[pagenumbers]{cvpr} % To force page numbers, e.g. for an arXiv version
\usepackage{orcidlink}
\usepackage{booktabs}
\usepackage{multirow}
\usepackage{footnote}
\usepackage{xr}
\usepackage{xspace}
\usepackage{caption}
\usepackage{times}
\usepackage{epsfig}
\usepackage{graphicx}
\usepackage{amsmath}
\usepackage{amssymb}
\usepackage{comment}

\usepackage{color, colortbl}

\newcommand{\cameramodel}{HumanFoV\xspace}
\newcommand{\hpsmodel}{CameraHMR\xspace}
\newcommand{\camsmplify}{CamSMPLify\xspace}

\newcommand{\mesh}{\mathcal{M}}
\definecolor{cvprblue}{rgb}{0.21,0.49,0.74}
\title{CameraHMR: Aligning People with Perspective }

\author{
    \begin{tabular}{c c}
        Priyanka Patel\textsuperscript{1} & Michael J. Black \\
        Meshcapade, Germany & Max Planck Institute for Intelligent Systems \\
        \texttt{priyanka@meshcapade.com} & \texttt{black@tuebingen.mpg.de} \\
    \end{tabular}
}

\begin{document}
\twocolumn[
{
    \renewcommand\twocolumn[1][]{#1}
   \maketitle
   \thispagestyle{empty}
    \vspace{-0.1in}
    \centering
    \begin{minipage}{1.00\textwidth}
\centerline{\includegraphics[width=1.0\textwidth]{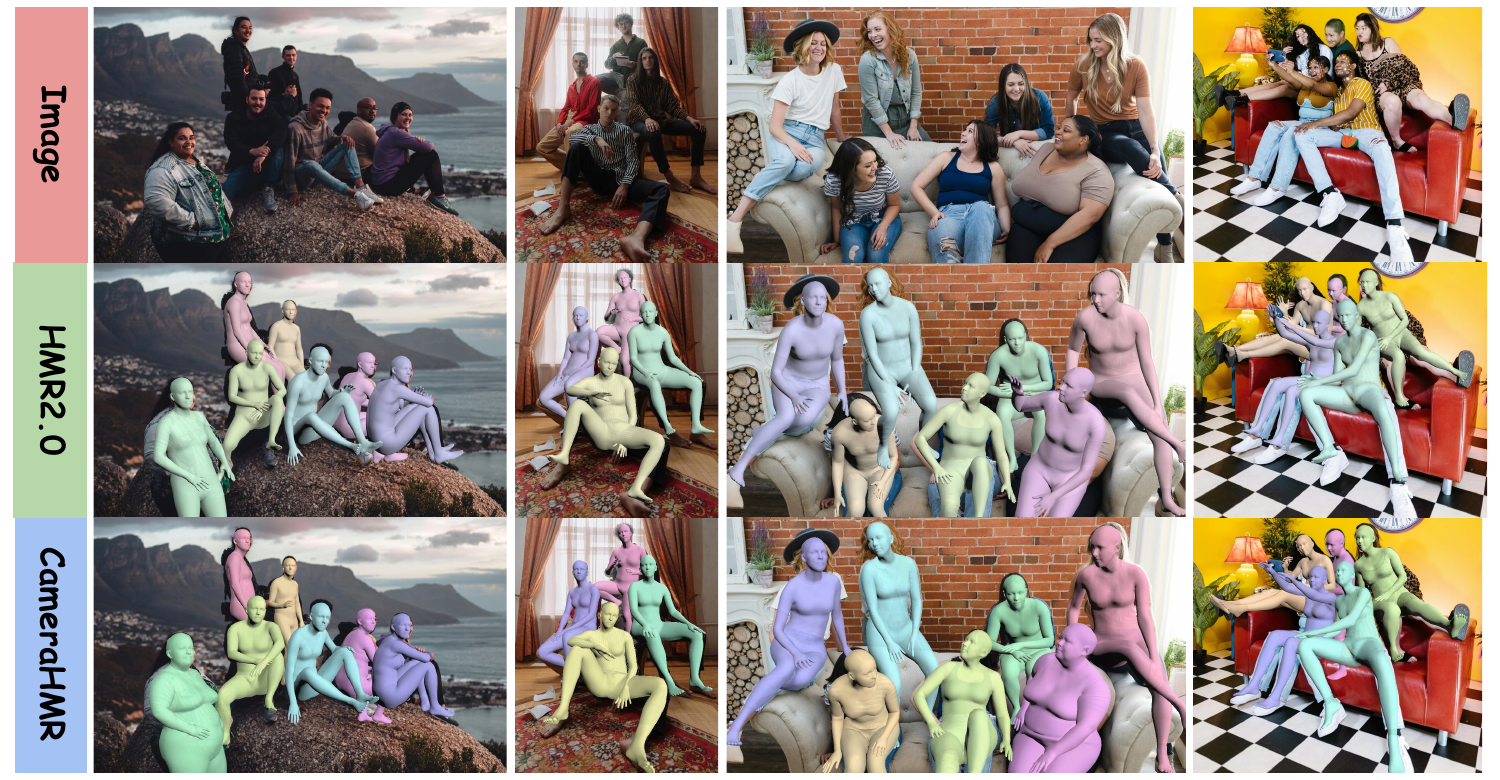}}
\vspace{-0.1in}
\end{minipage}
    \captionof{figure}{{\bf Putting people in perspective.}
    In contrast to common methods like HMR2.0, {\bf \hpsmodel} estimates 3D human shape and pose using a \textit{perspective camera} by leveraging a learned regressor, \textbf{\cameramodel}, to estimate the appropriate camera intrinsics. Note how this improves the estimated pose when there is strong \textit{foreshortening}. Our approach exploits new pseudo ground-truth data and a new dense surface keypoint detector that improve body shape estimation; this is particularly visible for the heavier people in the images. \hpsmodel defines the new state-of-the-art for 3D human pose and shape accuracy from a single image.
    }
	\label{fig:teaser}
     \vspace{2ex}

}
]
\footnotetext[1]{This work was done when PP was at MPI-IS.}
\begin{abstract}
%\vspace{-0.1in}
We address the challenge of accurate 3D human pose and shape estimation from monocular images. The key to accuracy and robustness lies in high-quality training data. Existing training datasets containing real images with pseudo ground truth (pGT) use SMPLify to fit SMPL to sparse 2D joint locations, assuming a simplified camera with default intrinsics. We make two contributions that improve pGT accuracy.
First, to estimate camera intrinsics, we develop a field-of-view prediction model (\cameramodel) trained on a dataset of images containing people. We use the estimated intrinsics to enhance the 4D-Humans dataset by incorporating a full perspective camera model during SMPLify fitting.
Second, 2D joints provide limited constraints on 3D body shape, resulting in average-looking bodies. To address this, we use the BEDLAM dataset to train a dense surface keypoint detector. We apply this detector to the 4D-Humans dataset and modify SMPLify to fit the detected keypoints, resulting in significantly more realistic body shapes.
Finally, we upgrade the HMR2.0 architecture to include the estimated camera parameters. 
We iterate model training and SMPLify fitting initialized with the previously trained model. 
This leads to more accurate pGT and a new model, \hpsmodel, with state-of-the-art accuracy. 
%The final \hpsmodel model trained with more accurate pGT achieves state-of-the-art 3D accuracy. 
%This leads to more accurate pGT and \hpsmodel trained on this data achieves state-of-the-art 3D accuracy. 
Code and pGT is available for research purposes.
\end{abstract}
    
\section{Introduction}
The field of monocular 3D human pose and shape (HPS) estimation has advanced rapidly. 
Updated architectures, stronger backbones, and more extensive training data have all led to improvements in robustness and accuracy.
We argue that a key remaining source of error lies in the fact that many HPS methods use a simplified weak-perspective camera model.
We describe how the wrong camera model introduces error and we propose a solution.
Specifically, we collect a dataset of images of people with varied field of view (FoV) and train a network to directly predict FoV from pixels.
We then leverage this predicted FoV in training and show how this leads to state-of-the-art accuracy.

%problem
Recent HPS methods, such as HMR2.0~\cite{hmr2}, achieve notable 2D alignment by leveraging large-scale real image datasets for training. However, this success in 2D alignment comes at the cost of reduced 3D accuracy as described in \cite{tokenhmr}. 
The core issue lies in the fact that these large-scale real image datasets frequently lack camera intrinsic parameters. 
Training involves first creating 3D pseudo ground truth (pGT) data by fitting a parametric 3D body model like SMPL \cite{smpl} to 2D features such as keypoints.
This fitting process typically uses a weak perspective camera model or default camera intrinsics.
When the camera model is wrong, fitting 2D keypoints accurately forces the 3D pose to be wrong.
Consequently, methods trained on these pGT datasets learn to replicate the 3D errors. 

To achieve both accurate 2D alignment and  3D poses, it is crucial to use the correct camera intrinsics in creating the pGT. 
% problem
Unfortunately, estimating intrinsics from a single image is challenging.
While there are many state-of-the-art approaches for camera calibration from monocular images \cite{ctrlc, perspectivefields, wildcamera}, they are trained on datasets such as Google Street View~\cite{gsv} and SUN360~\cite{sun360}.
Such datasets focus on outdoor or indoor scenes rather than people. 
Methods trained on datasets containing panoramic images of streets, natural landscapes, urban scenes, indoor settings, etc., do not work well on images of people.  
%As a result, they perform poorly on images with people and hence are not robust enough to be used by HPS methods. 
On the other hand methods trained on synthetic data like SPEC-camcalib~\cite{Kocabas_SPEC_2021} do not generalize  well to in-the-wild data. This highlights the need for a robust camera calibration model for images containing people to achieve accurate 3D human pose and shape estimation.

To address this problem, we collect a dataset of about 500K images predominantly comprising  people, to train a field of view (FoV) prediction model. 
The human body provides useful information for camera estimation. 
While it would be going too far to call the body a ``calibration object," bodies have highly regular proportions and a limited range of heights.
When projected into images, this regular structure systematically varies with focal length and perspective projection.
To exploit this fact, we train a direct FoV regressor, \textbf{\cameramodel}, which generalizes well on benchmarks featuring humans compared to other state-of-the-art (SOTA) camera calibration methods. 
\cameramodel can be directly incorporated into HPS methods that use a full perspective camera model, enabling accurate 3D reconstruction. 
Using an accurate camera not only helps HPS regressors infer the 3D location of the people in camera space but it also improves alignment of the inferred body with image features, especially for wide angle images and extreme viewing angles.

While incorporating a more accurate camera model into HPS methods is important, we still need high-quality training data that is as diverse as possible.
To that end, we use \cameramodel to improve the real-image pGT in the 4DHumans dataset that is used to train HMR2.0~\cite{hmr2}.
The original dataset uses SMPLify~\cite{smplify} to fit SMPL to 2D keypoints under a weak-perspective assumption.
Instead, we use a full perspective camera model in SMPLify and exploit \cameramodel to estimate the FoV of the training images. 

Additionally, the original dataset is created by fitting SMPL to only 17 sparse 2D joints; these lack the detail necessary for accurate 3D shape reconstruction. 
To improve this, we train a keypoint detector (\textbf{DenseKP}) on the BEDLAM~\cite{bedlam} dataset to estimate {\em 138 dense surface keypoints}.
We modify SMPLify to use these together with the original 17 2D joints.
This results in significantly more realistic body shapes.
Qualitatively, the improved camera model and dense keypoints lead to good 2D image alignment and more plausible 3D pGT compared to original dataset (Fig.~\ref{fig:dataset-comp}).

With this, we generate high-quality 3D pGT for a large-scale real image dataset comprising approximately 3.2 million cropped images. 
Importantly, the dataset includes the camera intrinsics estimated by \cameramodel; these are crucial for HPS methods. We further modify the HMR2.0 architecture to incorporate camera parameters from \cameramodel in training. 
We iterate training this new \textbf{\hpsmodel} model and refining the pGT with SMPLify initialized with the previously trained model.
%Incorporating the predicted camera intrinsics from our \cameramodel into the training our new \textbf{\hpsmodel} model 
This significantly improves performance, with \hpsmodel achieving state-of-the-art accuracy on multiple HPS benchmarks.
See Fig.~\ref{fig:teaser}.

In summary, we (1) collect a dataset of varied images of humans with known FoV, (2) using this dataset, we train \cameramodel to regress FoV from images of people, (3) update SMPLify with a full perspective model that uses the \cameramodel output,  (4) introduce a dense surface keypoint regressor and incorporate these keypoints into SMPLify,
(5) improve the 4DHumans training set using the new version of SMPLify, (6) incorporate the FoV estimation in HMR2.0, (7) train a new model, \hpsmodel, with SOTA accuracy.
Code and pGT dataset is available for research purposes.

\section{Related Work}
\label{sec:relatedwork}

\noindent\textbf{3D Human Pose and Shape Regression.}
The field of mononcular 3D human pose and shape (HPS) estimation has made rapid advances. 
The improvement of the backbone has played an important role, beginning with   ResNet architectures pre-trained on the ImageNet dataset~\cite{Kanazawa2018_hmr,spin,eft}, then the HRNet architecture pre-trained on the COCO dataset~\cite{hybrik,pymaf,pixie,expose},  and more recently  Transformer-based models~\cite{hmr2,refit,tokenhmr}.
These changes have led to significant improvements in accuracy on standard benchmarks.

HMR~\cite{Kanazawa2018_hmr} introduced a simplified weak perspective camera model to facilitate training with pseudo ground truth datasets. 
The availability of increasingly accurate 3D ground truth datasets, enables methods to be trained using the more complex full perspective camera model~\cite{beyondwp,li2022cliff,bedlam}. 
This evolution in camera modeling and training backbone has contributed to improvements in the accuracy of 3D pose and shape estimations. Despite advancements in achieving accurate 3D pose estimates, aligning these poses accurately with 2D image features remains challenging. 
This issue has been recently highlighted by TokenHMR~\cite{tokenhmr}, which attributes the misalignment to the use of incorrect camera models during prediction. 
Even methods that incorporate a full perspective camera model during training~\cite{li2022cliff,refit,beyondwp}, encounter alignment issues due to the absence of camera intrinsics during inference, leading to inaccurate projection into 2D. 

\noindent\textbf{Regressing Camera Intrinsics.}
Several approaches have been developed to regress camera intrinsics from monocular images~\cite{wildcamera,perspectivefields,ctrlc,spec}. However, these methods are often trained on datasets focused on indoor~\cite{sun3d,nyuv2}, driving~\cite{gsv,kitti}, or object-centric~\cite{objectron} images, which typically feature consistent vanishing points and geometric cues. 
These images are quite different from images of people.
Models trained on indoor and outdoor scenes struggle when presented with  images of people, particularly portrait images where vanishing points are often unclear or absent.
 We argue that the human body itself offers essential cues for camera parameter estimation. To leverage this, we train our model on a dataset composed predominantly of images featuring people. Our experiments demonstrate that this approach significantly enhances the generalization of camera estimation across various human-centric benchmarks.

\section{Method}

\begin{figure*}
\centerline{\includegraphics[width=1\textwidth]{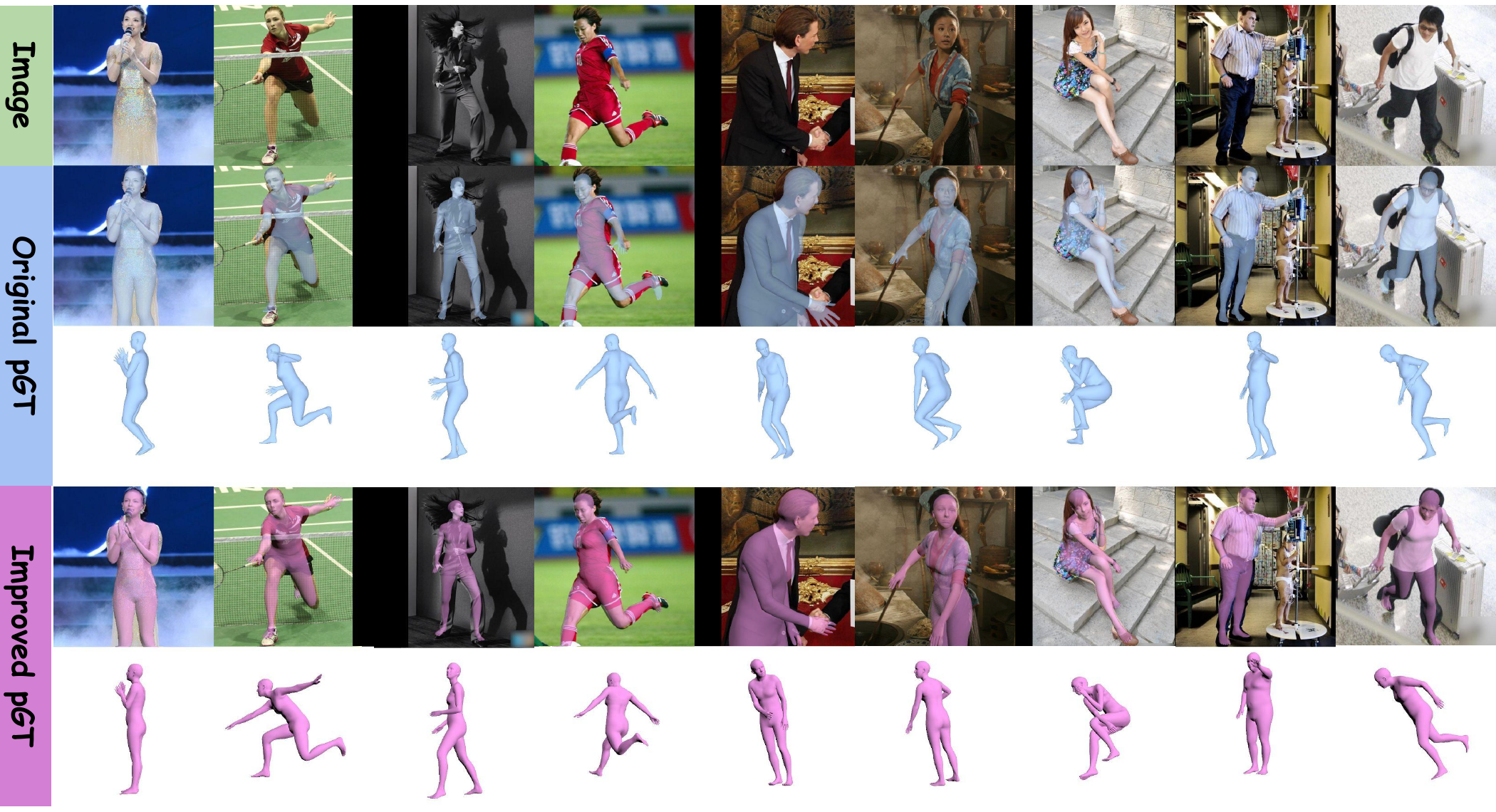}}
\vspace{-0.15in}
\caption{\textbf{Pseudo-Ground-Truth (pGT) training data.} 
Row 1: example images from the 4DHumans dataset.
Rows 2 and 3: original pGT overlaid and viewed from a different perspective.
Rows 4 and 5: our improved pGT using \camsmplify.
Note that our approach reduces the bias towards bent knees (columns 1, 5, 6), improves 3D pose and image alignment when there is foreshortening (Column 2, 4, 7, 9), and estimates more realistic body shape (columns 1, 3, 7, 8).
}
\label{fig:dataset-comp}
\end{figure*}

In this section, we detail our approach for estimating camera intrinsics from images captured in uncontrolled settings. 
We further present enhancements to the HMR2.0 architecture by incorporating bounding box and camera parameter tokens into the vision transformer and adopting a perspective camera model for projection, in contrast to the previously used weak perspective model. We designate the two models as \cameramodel and \hpsmodel. Furthermore, we describe the generation of improved pseudo-ground truth (pGT) by using a modified SMPLify fitting process, \textbf{\camsmplify}. 

\subsection{Preliminaries}

We use a simple perspective camera model for our experiments. In this model, a 3D point $(X, Y, Z)$ in camera coordinate space is projected onto the image plane, at a point $(x,y)$, using the camera intrinsic matrix  parameterized by focal length $f$ (assuming $f_x = f_y$) and principal point, $(c_x, c_y)$ in pixels coordinates. We simplify the model by assuming no radial distortion and that the principal point is the center of the image. 
For \hpsmodel, we assume the body is always predicted in camera space, i.e.~the rotation of the camera is $R=I$. 

As is common practice~\cite{wildcamera,ctrlc,spec}, we estimate the field of view from images instead of the focal length. The reason for this is that different focal lengths can produce images with same field of view if the camera sensor size is different. For example, a 50mm lens on a full-frame camera (35mm sensor) has a wider field of view than a 50mm lens with a crop sensor. While the focal length is an important characteristic of a lens, the field of view is a more direct measure of what will be captured in an image. Hence, we use vertical field of view $\upsilon$ as the primary output of \cameramodel. The focal length $f_y$ used in the camera intrinsics can be derived from $\upsilon$ using the image height $H$.
\begin{equation}
f_y  = \frac{H}{2 \cdot \tan\left(\frac{\upsilon}{2}\right)} .
\label{eq: fl_from_fov}
\end{equation}

To represent the 3D human, we use the SMPL parametric human body model \cite{smpl}  controlled by parameters $(\theta, \beta)$, where $\theta \in \mathbb{R}^{72}$ represents pose and $\beta \in \mathbb{R}^{10}$ represents identity shape. 
SMPL is a function that outputs a body mesh, $\mesh$  with vertices ${V} \in \mathbb{R}^{6890\times3}$. The 3D joints  ${J_{3d}} \in \mathbb{R}^{K\times3}$ with $K$ joints are obtained using a pre-trained joint regressor. We use a gender-neutral SMPL model with 10 shape components. We use a cropped bounding box of size $256\times256$ as input to our models.

\subsection{\cameramodel}

Given an image $I \in \mathbb{R}^{W\times H\times3}$ where $W$ and $H$ are the width and height of the original image respectively, we preprocess it by resizing it to a square resolution of $256 \times 256$ pixels. To achieve this, we resize the longer side to 256 and zero-pad  the smaller side to maintain the aspect ratio. This preprocessing ensures uniform input dimensions for the network while preserving the aspect ratio integrity of the original image, which is important for field of view estimation. 
We train a deep neural network architecture with HRNet~\cite{hrnet} as the backbone and an MLP head for direct estimation of the vertical field of view $\upsilon_{\text{pred}}$. The HRNet backbone is pretrained on ImageNet~\cite{imagenet}.  Based on insights from previous work ~\cite{spec,beyondwp}, underestimating the FoV has less negative impact on reconstructed 3D poses compared to overestimating the FoV. Therefore, an asymmetric loss function is incorporated to penalize overestimation more heavily than underestimation. We define a loss, $L_{\upsilon}$, on the vertical field of view in radians as:
\begin{equation}
L_{\upsilon} = 
\begin{cases} 
    3 \left\lVert \upsilon_{\text{gt}} - \upsilon_{\text{pred}} \right\rVert_{2}^{2} & \text{if } \upsilon_{\text{pred}} > \upsilon_{\text{gt}} \\
    \left\lVert \upsilon_{\text{gt}} - \upsilon_{\text{pred}} \right\rVert_{2}^{2} & \text{if } \upsilon_{\text{pred}} \leq \upsilon_{\text{gt}}.
\end{cases}
\label{eq:fov_loss}
\end{equation}

The training set for \cameramodel is described in Sec.~\ref{sec:HumanFoV}.
We train \cameramodel for around 16 epochs with a batch size of 64 and learning rate of $5 \times 10^{-5}$. We use an Adam
optimizer~\cite{adam} with no weight decay.  We use different data augmentation techniques during training to ensure the model's robustness. This includes  center-cropping of images to generate different aspect ratios. This augmentation helps the model become robust against variations in image cropping during inference. Images are also randomly flipped horizontally with a probability of 0.2, providing additional diversity to the training dataset.

\subsection{\hpsmodel}
HPS methods have evolved to use progressively more powerful backbones from ResNet~\cite{resnet} to HRNet~\cite{hrnet}, and most recently ViT~\cite{vit}, resulting in improved performance.
Here we use a ViTPose~\cite{vitpose} backbone pretrained on COCO~\cite{coco} to extract features from cropped images. Specifically, we adopt the HMR2.0 architecture, which employs a ViT backbone, and modify it to support training with a full perspective camera instead of a weak perspective camera. 

The ViT backbone processes images by dividing them into patches, converting these patches into feature embeddings known as tokens, and utilizing self-attention mechanisms to capture the relationships among them. Along with the image tokens, we also provide bounding box information of the cropped region and the  focal length as tokens. We follow CLIFF~\cite{li2022cliff} and compute the bounding box token $T_{\text{bbox}}$, using the bounding box center $c_x, c_y$, scale $s$, and the focal length $f$ of the full image. 
\begin{equation}
    \mathbf{T}_{\text{bbox}} = \left( \frac{c_x}{f}, \frac{c_y}{f}, \frac{s}{f} \right)
\end{equation}
The ground truth focal length is known during training and predicted using our \cameramodel during inference. 

The decoder in our modified architecture cross-attends to both image-derived tokens and the supplementary bounding box and focal length tokens. This approach enables the decoder to generate features essential for accurately regressing 3D rotations and human mesh parameters while accommodating camera perspective. We explain the losses used in training the model in  Sup.~Mat.
\begin{figure}
\centerline{\includegraphics[width=1.0\columnwidth]{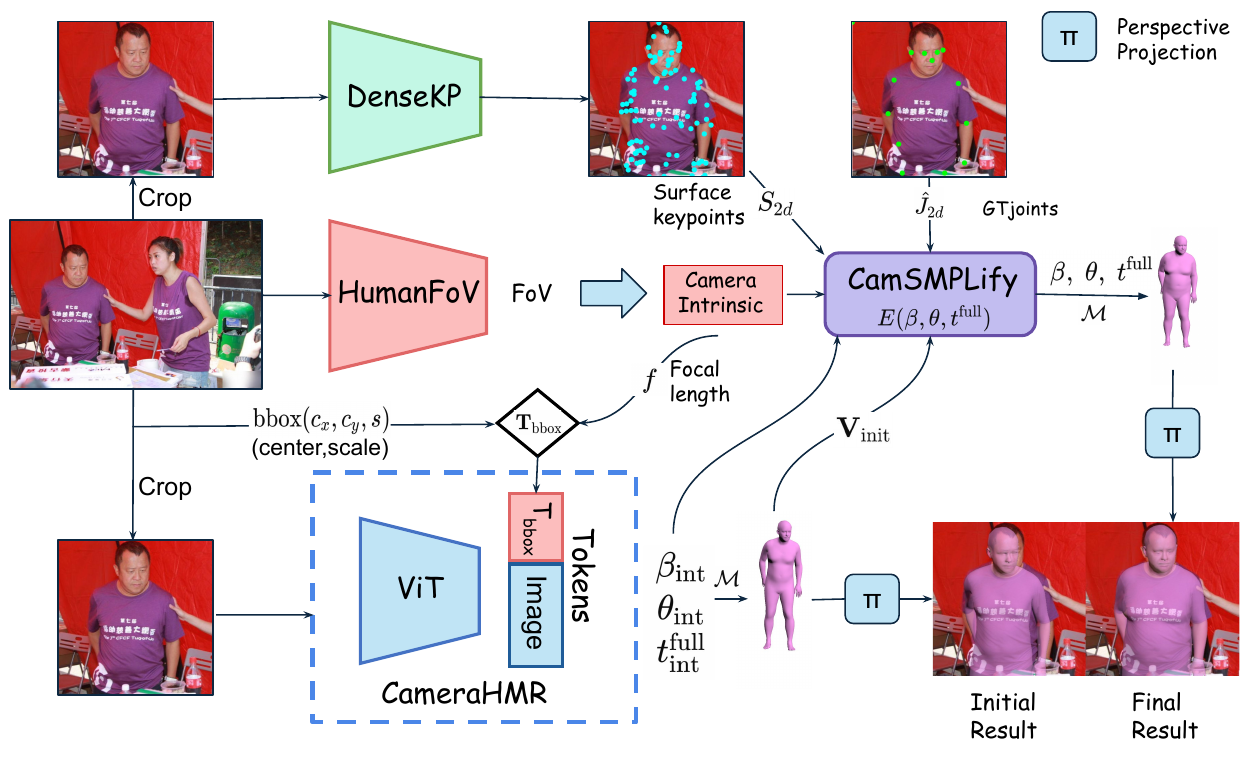}}
\vspace{-0.1in}
\caption{\textbf{Overview of CamSMPLify}: The DenseKP module processes cropped images to produce dense surface keypoints, while the \cameramodel module uses full images to estimate camera intrinsics. 
The output from these are used by \camsmplify  to optimize the SMPL model parameters, $\beta$, $\theta$, and the global translation $t^{\text{full}}$.
Our iterative training strategy 
starts with initial estimates, $\mathrm{V}_{init}$ from \hpsmodel, which are used to regularize the \camsmplify estimates.  
\hpsmodel is then iteratively refined based on the improved pGT from \camsmplify.}
 \vspace{-0.15in}
	\label{fig:smplify_diag}
\end{figure}

\subsection{\camsmplify}
To train \hpsmodel, we modify the original 4DHumans~\cite{hmr2} dataset, upgrading it from a weak perspective to a full perspective camera format. This dataset includes images from the
InstaVariety~\cite{insta}, COCO~\cite{coco}, MPII~\cite{mpii}, AI Challenger~\cite{aic}
%InstaVariety~\cite{insta}, AI Challenger~\cite{aic} 
and AVA~\cite{ava} datasets.
Note that AVA~\cite{ava} contains many movie clips where the aspect ratio is stretched horizontally (e.g.~from 4:3 to 16:9), violating the assumption that $f_x = f_y$. These clips are hard to detect automatically so we exclude AVA from our improved pGT.
We make several improvements to the fitting process to enhance the dataset, including better initialization and priors, more accurate camera intrinsics, dense surface 2D keypoints, and multiple fitting iterations. The overview is shown in Fig.~\ref{fig:smplify_diag}. This comprehensive approach results in better pseudo ground truth, significantly improving both 3D pose and shape accuracy, as well as 2D alignment as shown in Fig.~\ref{fig:dataset-comp}. In the following section, we explain each step in greater detail.

\noindent\textbf{Camera Intrinsics.}
One major challenge with in-the-wild image datasets is the absence of ground truth intrinsic camera parameters, which makes it difficult to accurately fit a 3D body to the 2D keypoints while ensuring plausible 3D poses and precise 2D alignment. To address this issue, we employ our  \cameramodel model to estimate the vertical field of view $\upsilon$ for all images in the 4DHumans dataset. The corresponding focal length $f$ is then calculated from $\upsilon$ and image height $H$ using Eq.~\ref{eq: fl_from_fov}. This approach enables us to infer the necessary camera intrinsics for the dataset, enabling us to project the 3D joints using a full perspective camera in contrast to the weak perspective camera used in the original dataset. As a result, we achieve accurate 2D alignments without compromising 3D pose accuracy during the fitting.

\noindent\textbf{Surface Keypoints.}
Another challenge that we face is that the original 4DHumans dataset is annotated with 17 sparse 2D body joints. This sparse annotation lacks sufficient detail to accurately reconstruct the 3D body shape.
To address this, we train a dense surface keypoint detector, DenseKP, using the synthetic datasets BEDLAM and AGORA, which offer diverse body shapes and precise ground truth annotations. We use ViTPose~\cite{vitpose} pretrained on COCO to extract features from the cropped images. The model takes a centered crop of the person, resized to $256 \times 256$ pixels, as input and outputs 2D dense surface keypoints $S_{2d} \in \mathbb{R}^{138 \times 2}$. 
\textcolor{black}{To generate the 138 surface keypoint ground truth labels, we down-sample the SMPL ground truth vertices and project them onto the image. The down-sampling approach, adapted from COMA~\cite{coma}, focuses on sampling vertices in high-curvature regions, which effectively preserves the body shape and key structural details.}
To train the model we use an $L_2$ loss on ground truth $\hat{S}_{2d}$ and predicted keypoints $S_{2d}$. 
We use this model to generate 138 dense keypoints for all images in the 4DHumans dataset. 
We modify the fitting by combining the 17 provided joints with the estimated 138 dense surface keypoints. 
This, along with improved camera intrinsics, helps achieve better shape alignment in 2D during projection.  

\noindent\textbf{Initialization.}
Like with any optimization-based method, the choice of initialization can significantly affect the performance of the fitting process, as poor initialization can lead to suboptimal convergence. To ensure a robust 3D initialization, we train our \hpsmodel on the BEDLAM and AGORA dataset, which provides accurate ground truth annotations, including camera intrinsics.
Using this pre-trained model, we generate initial 3D pose $\theta_{\text{int}}$ and shape  $\beta_{\text{int}}$ predictions for the 4DHumans dataset, resulting in bodies with relatively accurate pose and shape. Additional refinement through surface keypoint and joint fitting improve both the body shape and pose, leading to even more precise results. We also predict the initial global translation of the mesh $t^{\text{full}}_{\text{int}}$ relative to the optical center of the full image. 
Note that instead of the standard pose prior used in SMPLify \cite{smplify}, we regularize the solution to this initial prediction.

\noindent\textbf{Losses.}
We optimize the SMPL model parameters $\beta$ and $\theta$ to match the ground truth 2D body joints $\hat{J}_{2d}$ and 2D surface keypoints $\hat{S}_{2d}$ while also optimizing the global translation of the mesh $t^{\text{full}}$. The model parameters are initialized with $\beta_{\text{int}}$ and $\theta_{\text{int}}$, as determined during the initialization stage using \hpsmodel. The output vertices $V_{\text{int}}$ from the model $M(\beta_{\text{int}}, \theta_{\text{int}})$ serve as the prior in the regularization process. $J_{2d}$ and $S_{2d}$ are  obtained from the 3D joints $J_{3d}$ and surface keypoints $S_{3d}$ using $\Pi(J_{3d} + t^{\text{full}})$, where $\Pi$ represents the perspective projection with camera intrinsics obtained from \cameramodel. The $S_{3d}$ are derived from the SMPL vertices $V$ using a downsampling matrix similar to BEDLAM~\cite{bedlam}.
Specifically the optimization minimizes the following objective function:
\begin{equation}
E(\beta, \theta, t^{\text{full}}) = \lambda_{S_{2d}} E_{S_{2d}} + \lambda_{J_{2d}} E_{J_{2d}} + E_\text{reg}
\end{equation}
\begin{equation}
E_\text{reg} = \lambda_{\beta} \|\beta\|_2^2 + \lambda_{\text{int}} \| V - V_{\text{int}} \|_2^2,
    \label{main:eq:reg}
\end{equation}
where $E_{J_{2d}}$ and $E_{S_{2d}}$ are $L_2$ losses on 2D joints and surface keypoints, respectively. The $\lambda$ terms denote the weights for each component of the objective function.
We apply a threshold value $\tau$ to filter out results with $E > \tau$, thereby excluding pseudo ground truth samples with high convergence errors. For further details, refer to the Sup.~Mat.

\noindent\textbf{Iteration.}
We run the whole process for multiple iterations of refinement to ensure the pseudo ground truth is of high quality. This is similar in spirit to SPIN \cite{spin}.
The $\theta$ and $\beta$ parameters for \camsmplify fitting are initialized using version \textbf{v1} of the \hpsmodel trained on the BEDLAM dataset. After applying the filtering criteria based on the threshold $\tau$ in \camsmplify fitting we obtain approximately 2.8 million crops out of 4 million crops from 4DHumans dataset. These crops, together with the BEDLAM dataset, are then utilized to train an enhanced version, \textbf{v2}, of \hpsmodel. 
We further iterate and employ \textbf{v2} to generate improved initializations for the 4D Humans dataset, followed by another round of \camsmplify fitting.  This improved initialization substantially improves convergence, leading to a more accurate fitting with lower error. Applying the same filtering criteria, we are able to further expand our dataset to around 3.2 million high-quality annotations.

\section{Datasets}

\subsection{\cameramodel}
\label{sec:HumanFoV}
To train the \cameramodel model we use around 500K images collected from Flickr~\cite{flickr}. To get human-centeric data, we filter Flickr with keywords such as people, man, woman, kid, human, crowd etc. 
We use the Flickr API to download only the images that have associated EXIF information, which usually contains the focal length in mm. 
To calculate the vertical field of view $\upsilon$ from the focal length $\mathsf{f}$ in mm, we need to know the sensor height $sh$. Vertical FoV $\upsilon$ can be calculated as
\begin{equation}
     \upsilon = 2 \cdot \arctan\left(\frac{sh}{2 \cdot \mathsf{f}}\right).
\end{equation}
We use the field \textit{FocalLengthIn35mmFormat} from the EXIF, which contains the focal length corresponding to a 36mm wide and 24mm high sensor. This allows us to use a sensor height of 24mm directly in our calculations. Note that if the aspect ratio of the image is less than 1 (i.e.~portrait mode), we calculate $\upsilon$ using the sensor width instead of sensor height. Please refer to the Sup.~Mat.~to see the distribution of focal lengths in the dataset.

Note that the standard aspect ratios for images captured from a phone or camera are in the range of 16:9, 4:3, 3:2, 1:1, 5:4. If the aspect ratio of the image collected from Flickr is outside this range, we assume that the image is cropped. Cropped images, especially those not centered, could introduce inconsistencies with the camera model used and  might confuse the neural network. We filter such images from the training data. Although we do not use cropped images directly in our training data, we extensively apply crop augmentation during training to ensure model's robustness.

To evaluate our \cameramodel model, unlike previous methods, we focus on benchmarks containing images of people. Consequently we use test-set images from SPEC~\cite{spec}, 3DPW~\cite{3dpw} and EMDB~\cite{emdb}, which all provide camera intrinsics. 
We also create a test set of around 10K images from Flickr that are similar to training set. 
We also include a scene from the BEDLAM~\cite{bedlam} dataset, BEDLAM-Z, which contains a wide range of focal lengths because the camera is zooming from 28 to 80mm. This provides a more varied distribution over the intrinsics in the test set.

\subsection{\hpsmodel}

For training \hpsmodel along with the enhanced 4DHumans dataset (minus AVA) we also use the synthetic datasets AGORA~\cite{agora} and BEDLAM~\cite{bedlam}, which contain accurate ground truth camera information. 
The combination of all these datasets is called ``All" in Table \ref{hps}.
For evaluation we use the 3DPW~\cite{3dpw} EMDB~\cite{emdb} and RICH~\cite{rich} datasets. Additionally, we evaluate accuracy on the  SPEC test set~\cite{spec}, which features multiple off-center individuals and more varied camera perspectives. To evaluate 3D shape accuracy, we utilize the SSP-3D~\cite{ssp} dataset. We also evaluate 2D alignment accuracy on the COCO validation set and perform qualitative evaluation using images from the LSP~\cite{lsp} and MPII~\cite{mpii} test sets in 
Fig.~\ref{fig:baseline-comp}.

\subsection{Evaluation Metrics}
We follow previous work \cite{spec, hmr2} and evaluate  3D reconstruction accuracy using MPJPE (Mean Per Joint Position Error), PA-MPJPE (Procrustes Analysis Mean Per Joint Position Error), and PVE (Per Vertex Error), which measures the Euclidean distance (in mm) between predicted and actual 3D vertices and joints after aligning the pelvis. 
PVE is useful for evaluating body shape accuracy.
We also evaluate the 2D alignment using PCK (Percentage of Correct Keypoints) on COCO-val~\cite{coco}. PCK measures the accuracy of 2D keypoint detection by calculating the percentage of predicted keypoints within a specified distance threshold from the ground truth.
We use thresholds of \(0.05\) and \(0.1\), which correspond to \(5\%\) and \(10\%\) of the crop size, respectively.

\begin{table}
\resizebox{\columnwidth}{!}{
\begin{tabular}{lccccc}
\toprule
& Flickr-test & SPEC & 3DPW & EMDB & BEDLAM-Z \\ 
\midrule
Perspective Fields~\cite{perspectivefields} & 15.3 & {\it 8.0} & 14.0 & 12.8 & 18.0 \\
Ctrl-C~\cite{ctrlc} & 26.5 & 10.4 & {\it 5.6} & {\it 5.4} & 31.3\\
WildCamera~\cite{wildcamera} & {\it 10.9} & 17.8 & 8.2 & {\bf 2.3} & {\bf 4.8}\\
SPEC-camcalib~\cite{spec} & 14.0 & 14.3 & 8.8 & 5.9 & 14.2              \\
\cameramodel (Ours) & {\bf 7.3}  & {\bf 7.9} &  {\bf 5.0}  &5.7 & {\it 5.4}               \\
\bottomrule
\end{tabular}
}
\caption{Vertical field of view error in degrees. BEDLAM-Z stands for the ``zoom" sequence from BEDLAM used for testing (see text). Bold and italics correspond to best and second best respectively.}
\label{fov}
\end{table}

\begin{table}
\centering
{\scriptsize
\begin{tabular}{lccc}
    \toprule
     Method & PCK 0.05 $\uparrow$ & PCK 0.1 $\uparrow$\\
    \midrule
    CLIFF~\cite{li2022cliff}  & 0.66 & 0.84\\
    BEDLAM-CLIFF~\cite{bedlam} & 0.62 & 0.80\\
    HMR2.0a~\cite{hmr2} & 0.79 & 0.94\\
    HMR2.0b~\cite{hmr2} & \textbf{0.86} & \textbf{0.96}\\
    TokenHMR~\cite{tokenhmr} & 0.80 & \textit{0.95} \\
    ReFit~\cite{refit} & 0.74 &  0.84 \\
    CameraHMR (Ours) & \textit{0.84} & 0.94 \\
    
    \bottomrule
  \end{tabular}
  }
  \caption{PCK on COCO-val dataset measures 2D alignment accuracy. 
  Bold: most accurate. Italics: second most.}
  
\label{pck}

\end{table}

\section{Experiments}

\subsection{Comparision to SOTA}
\noindent\textbf{\cameramodel.}
As shown in Table~\ref{fov},  \cameramodel generalizes well across all benchmarks with diverse fields of view. While WildCamera~\cite{wildcamera} excels on benchmarks with narrow to average field of view, it underperforms on those with wide field of view, such as SPEC. In contrast, our \cameramodel maintains consistent accuracy across all benchmarks.

\noindent\textbf{\hpsmodel.}
As shown in Table~\ref{hps}, \hpsmodel outperforms the baselines on all three benchmarks by a large margin. For a fair comparison, we categorize the methods based on the major datasets that were used in training. 
STD refers to standard datasets comprising Human3.6M~\cite{h36m}, COCO~\cite{coco}, MPII~\cite{mpii} and MPI-INF-3DHP~\cite{mpiinf3dhp} while 4DHumans comprises InstaVariety~\cite{insta}, COCO~\cite{coco}, MPII~\cite{mpii}, AI Challenger~\cite{aic} and  AVA~\cite{ava}. 
Note that \hpsmodel training never uses AVA.
Even when trained on similar datasets, \hpsmodel consistently outperforms all other baselines, demonstrating notable improvements particularly on the  EMDB and SPEC-SYN benchmarks, which feature a wide variety of cameras. 
Additionally, our estimated dense keypoints improve the pGT body shape accuracy and this translates into improved accuracy of \hpsmodel on the SSP-3D \cite{ssp} dataset; see Sup.~Mat. 
%and HBW \cite{shapy} datasets

\begin{figure*}
\centerline{\includegraphics[width=0.9\textwidth]{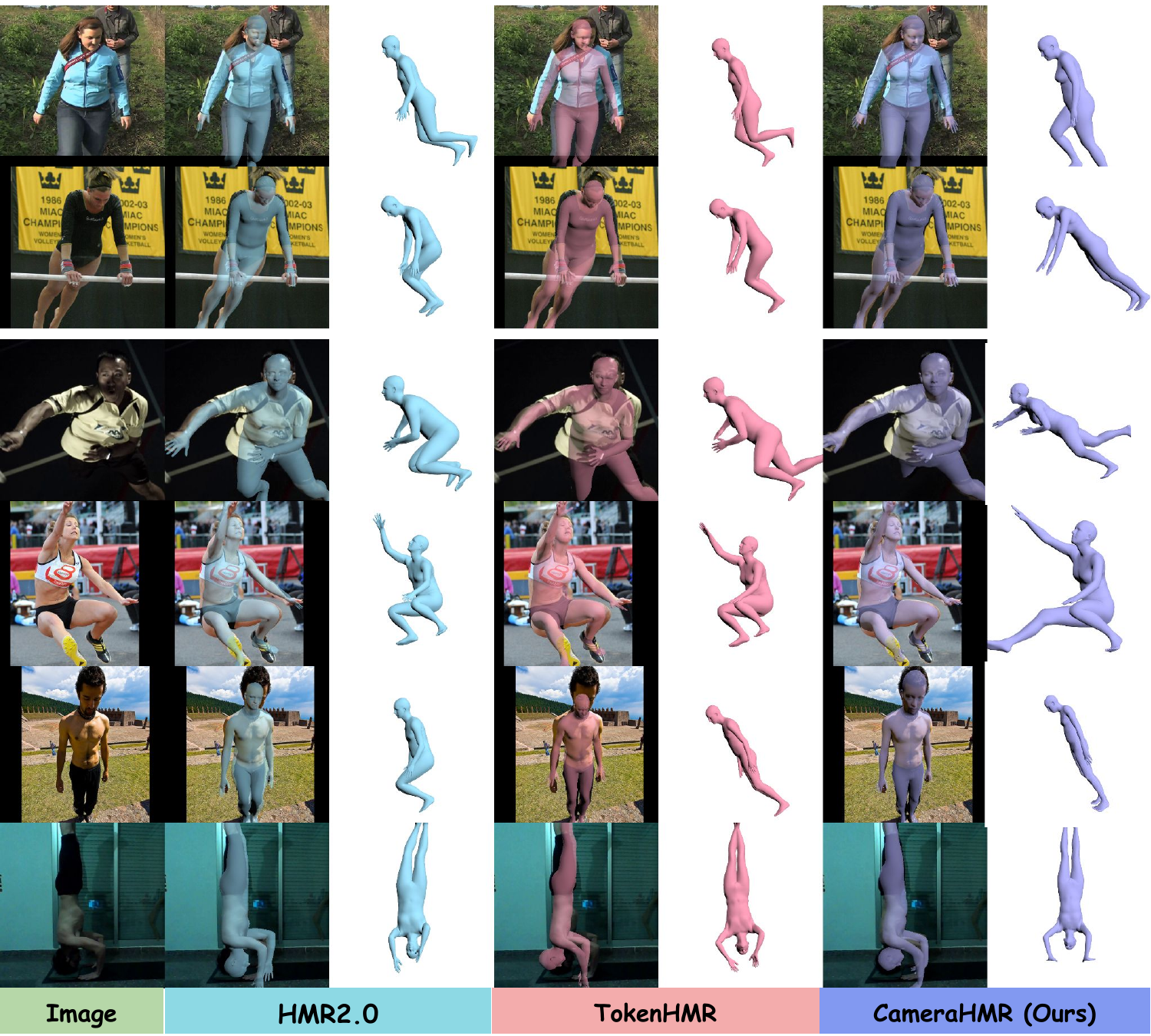}}
	\caption{\textbf{Qualitative results of different baselines on LSP~\cite{lsp} and MPII~\cite{mpii} test images}. \hpsmodel achieves better 3D pose and shape reconstruction while also achieving more accurate 2D alignment compared to other SOTA methods trained on comparable datasets. }
	\label{fig:baseline-comp}
\end{figure*}

Table~\ref{pck} also shows that \hpsmodel is either comparable to, or better than, other baselines in terms of 2D alignment on the COCO-val dataset.
\hpsmodel is nearly identical to HMR2.0b in terms of 2D keypoint alignment, while having significantly better 3D accuracy.
As shown qualitatively  in Fig.~\ref{fig:baseline-comp}, \hpsmodel achieves not only better 3D reconstruction, but significantly better 2D alignment, especially in cases of foreshortening or for people with non-average body shapes.

\begin{table}
\centering
{\scriptsize
  \begin{tabular}{lccc}
    \toprule
     Camera & EMDB~\cite{emdb}$\downarrow$ & RICH~\cite{rich}$\downarrow$ & SPEC-SYN~\cite{spec}$\downarrow$ \\
    \midrule
    fixed   & 89.3 & 64.8 & 138.7 \\
    default  & 82.7& 64.9 & 115.2 \\
    predicted   & \textbf{81.7}& \textbf{64.4}& \textbf{72.9} \\
    % ground truth  & & &  \\
    \bottomrule
  \end{tabular}
  }
  \caption{Per-vertex error (PVE) in mm for different focal length used during inference of \hpsmodel.}
\label{abl}
\end{table}

% FOV EVALUATION
% HPS EVALUATION
\begin{table*}
  \centering 
  \scriptsize
  \begin{tabular}{ll|ccc|ccc|ccc}
    \toprule
 &\multirow{2}{*}{Method}& \multicolumn{3}{c}{3DPW~\cite{3dpw}} & \multicolumn{3}{c}{EMDB~\cite{emdb}} & \multicolumn{3}{c}{SPEC-SYN~\cite{spec}} \\
    \cmidrule(lr){3-5}
    \cmidrule(lr){6-8}
    \cmidrule(lr){9-11}
     && PA-MPJPE $\downarrow$ & MPJPE $\downarrow$ & PVE$\downarrow$ & PA-MPJPE$\downarrow$  &MPJPE$\downarrow$ & PVE$\downarrow$  & PA-MPJPE $\downarrow$ & MPJPE$\downarrow$ & PVE$\downarrow$  \\
    \midrule
    \multirow{2}{*}{\rotatebox[origin=c]{90}{STD}} 
    &SPEC~\cite{spec} & 53.2& 96.5& 118.5 &  87.7 & 138.9 & 161.3 & 56.9 & 83.5 & 98.9    \\
    &CLIFF$^*$~\cite{li2022cliff} & 43.0& 69.0 & 81.2 & 68.3 & 103.5 & 123.7 & 55.8 & 128.5 & 139.0 \\
    &HMR2.0a$^*$~\cite{hmr2} & 44.4 & 69.8 & 82.2 & 61.5 & 97.8 & 120.0 & 55.8 & 133.3 & 153.0\\
    \midrule
    \multirow{5}{*}{\rotatebox[origin=c]{90}{BEDLAM}} 
    &TokenHMR~\cite{tokenhmr} &43.8 & 70.5 & 86.0 & 49.8 & 88.1 & 104.2 & 51.8 & 110.5 & 127.6\\
    &WHAM$^{\dagger*}$~\cite{wham} & 35.9 & 57.8 & 68.7 & 50.4 & 79.7 & 94.4 & - & - & - \\
    &ReFit$^*$~\cite{refit} & 38.2 & 57.6 & 67.6 & 55.5 & 91.7 & 106.2 & 51.3 & 103.6 & 116.3 \\
    &BEDLAM-CLIFF~\cite{bedlam} & 46.6 &72.0 & 85.0& 61.3 & 97.1 & 113.2& 55.6 & 109.9 & 124.6\\
    &\hpsmodel(Ours) & 40.0 & 62.3 & 74.8 & 45.4 & 82.7 & 97.0 & \textbf{31.8} & \textbf{58.9} & \textbf{70.0}\\
    \midrule
    % SPEC & 96.5& 118.5 &143.0& 166.0  & 410.1 & 81.0 & 95.5 &  1381.3\\
    % BEV & 78.5 & 92.3 & 106.4 & 123.6& 458.8 &114.7 & 131.0& 3827.7 \\
    % \hpsmodel (default cam) & \textbf{62.6}& \textbf{74.3} &\textbf{78.1} & \textbf{92.0} &  &94.3 &104.3 & 5207.6\\  
    \multirow{2}{*}{\rotatebox[origin=c]{90}{4DH}} 
    &HMR2.0b ~\cite{hmr2} & 54.3 & 81.3 & 93.1 & 79.2 & 118.5 & 140.6 & 67.6 & 150.7 & 172.9\\
    &\hpsmodel(Ours) & 38.7 & 62.7 & 73.4& 43.9& 73.2 & 85.6&  37.0 & 66.0 & 79.1\\
    \midrule
    \multirow{2}{*}{\rotatebox[origin=c]{90}{All}} 

    &\hpsmodel(Ours) & 38.5 & 62.1 & 72.9 & 43.7 & 73.0 & 85.4 &33.0 & 61.8 & 73.1 \\
    &\hpsmodel$^*$(Ours) & \textbf{35.1} & \textbf{56.0} & \textbf{65.9} & \textbf{43.3}& \textbf{70.3} &\textbf{81.7} & 32.9 & 61.8 & 72.9  \\

   % \hpsmodel (gt cam) & 62.6 & 75.2 & 85.3& 98.4& &  & & \\
    \bottomrule
  \end{tabular}
  \caption{\textbf{Reconstruction error comparison for HPS.} $^*$denotes method is finetuned on 3DPW training data. $^\dagger$denotes video based method. Datasets used in training; STD: standard datasets, 4DH: 4DHumans dataset, All: \text{BEDLAM} + \text{4DHumans dataset}. }
\label{hps}
\end{table*}
\subsection{Ablation}
To understand the effect of using the predicted camera intrinsics from our \cameramodel, we perform an ablation study in which we vary the focal length used during inference. We use a fixed focal length of 5000 pixels, the predicted focal length from  \cameramodel, and the default focal length that is use by other HPS methods \cite{li2022cliff, beyondwp}, defined as $\sqrt{w^2 + h^2}$
where \( w \) and \( h \) are the width and height of the full image respectively.
We evaluate per-vertex error on three different benchmarks: EMDB, RICH, and SPEC-SYN. As shown in Table~\ref{abl}, the impact of using the predicted focal length is modest on  EMDB and RICH. However, there is a significant improvement on the SPEC-SYN benchmark. EMDB and RICH largely contain centered individuals with fixed camera intrinsics, resulting in similar performance whether using predicted or default focal lengths. In contrast, SPEC-SYN includes varied camera intrinsics and off-center subjects,
resulting in foreshortening and perspective distortion.
In such cases, the benefits of using predicted focal length over the default focal length is significant.

\section{Conclusion}

In this work, we address the limitations of using an incorrect camera model in 3D human pose and shape estimation. By developing \cameramodel, a robust FoV predictor trained on a diverse human-centric dataset, we significantly enhance the accuracy of 3D human pose and shape estimation. Our integration of a full perspective model and dense surface keypoints into the SMPLify process improves the quality of pseudo ground truth data for in-the-wild images. Incorporating these advancements into the training of \hpsmodel results in state-of-the-art performance on various benchmarks, demonstrating the effectiveness of our approach in improving both 2D alignment and 3D reconstruction.

{\small \noindent\textbf{Disclosure.}
MJB has received research gift funds from Adobe, Intel, Nvidia, Meta/Facebook, and Amazon.  MJB has financial interests in Amazon and Meshcapade GmbH.  While MJB is a co-founder and Chief Scientist at Meshcapade, his research in this project was performed solely at, and funded solely by, the Max Planck Society.
}

{
    \small
    \bibliographystyle{ieeenat_fullname}
    \bibliography{main}
}
\clearpage
\clearpage
\maketitlesupplementary
\section{Focal length distribution}

\begin{figure}[t]
\centerline{\includegraphics[width=0.95\columnwidth]{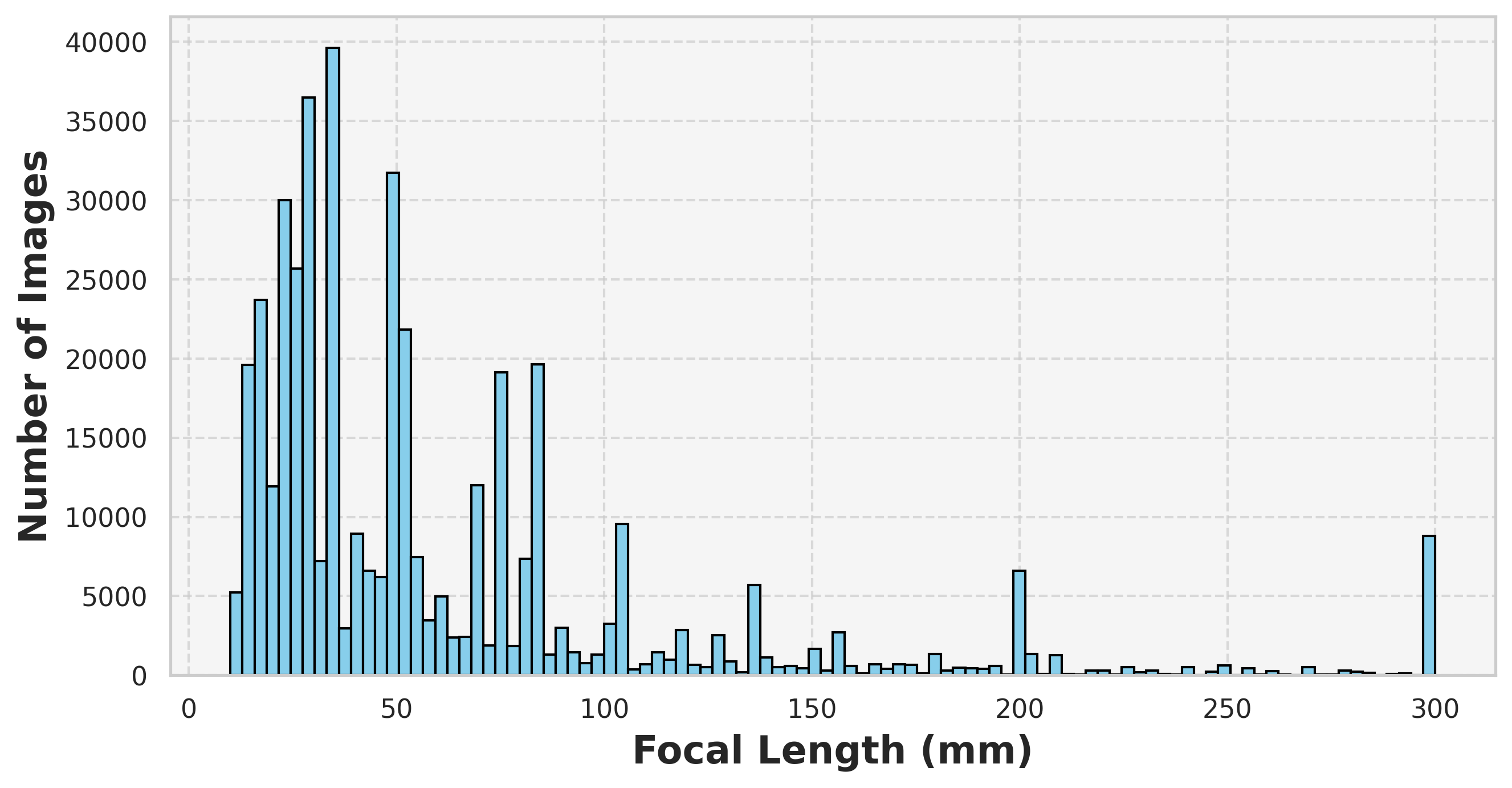}}
\vspace{-0.1in}
 \caption{Focal Length distribution of images used in training \cameramodel.}
 \vspace{-0.15in}
	\label{fig:fl_dist}
\end{figure}

We plot the distribution of focal lengths used in training HumanFoV model in Figure~\ref{fig:fl_dist}. The distribution shows notable peaks corresponding to the focal lengths of lenses most frequently used in photography, e.g.~24, 28, 35, 50, 85, 105, 135, 200, 300,  
\section{CameraHMR}
\subsection{Losses}
We use several loss functions to ensure accurate 3D human pose and shape estimation. We minimize the L2 norm distance between the ground truth 3D joint locations $\hat{\mathbf{J}}_{3d} \in \mathbb{R}^{44 \times 3}$ and the predicted 3D joint locations $\mathbf{J}_{3d} \in \mathbb{R}^{44 \times 3}$ centered around pelvis joint using $\mathcal{L}_{J_{3d}}$.

\begin{equation}
\mathcal{L}_{j3d} = \|\hat{\mathbf{J}}_{3d} - \mathbf{J}_{3d}\|_2^2
\end{equation}

We also penalize the deviation of the ground truth SMPL shape and pose parameters $\hat{\boldsymbol{\beta}}$ and $\hat{\boldsymbol{\theta}}$ from the predicted  parameters $\boldsymbol{\beta}$ and $\boldsymbol{\theta}$, respectively using $\mathcal{L}_{\text{SMPL}}$. The network predict $\boldsymbol{\theta}$ in a 6D rotation format~\cite{zhou}, which is later converted to a rotation matrix representation to facilitate loss calculation. This avoids the discontinuities and ambiguities associated with angle-based representations, such as Euler angles or quaternions.

\begin{equation}
\mathcal{L}_{\text{smpl}} = \|\hat{\boldsymbol{\beta}} - \boldsymbol{\beta}\|_2^2 + \|\hat{\boldsymbol{\theta}} - \boldsymbol{\theta}\|_2^2
\end{equation}

Since we have accurate body shape labels we also minimize the difference between the ground truth 3D body mesh vertices $\hat{\mathbf{V}}_{3d} \in \mathbb{R}^{6890 \times 3}$ and the predicted vertices $\mathbf{V}_{3d} \in \mathbb{R}^{6890 \times 3}$, using $\mathcal{L}_{v3d}$

\begin{equation}
\mathcal{L}_{v3d}  = \|\hat{\mathbf{V}}_{3d} - \mathbf{V}_{3d}\|_2^2
\end{equation}

For the 2D keypoint loss, we project ${\mathbf{J}}_{3d}$ and $\hat{\mathbf{J}}_{3d}$ onto the 2D image plane using the camera intrinsic matrix $\mathbf{K}$. For BEDLAM~\cite{bedlam} and AGORA~\cite{agora}, the ground truth intrinsic matrix $\mathbf{K}$ is provided. For 4DHumans dataset, where ground truth camera parameters are unavailable, we estimate $\mathbf{K}$ using our HumanFoV model. The 2D projection ${\mathbf{J}}_{2d}$ of the 3D joints is computed as $\Pi({\mathbf{J}}_{3d}; \mathbf{K})$ where $\Pi(\cdot)$ denotes the projection using the intrinsic matrix $\mathbf{K}$. 

Given that $ J_{2d} $ represents points in the original image coordinates where the horizontal and vertical coordinates $ x $ and $ y $ satisfy $ x \in [0, W] $ and $ y \in [0, H] $ respectively, we need to normalize these coordinates before calculating the loss. Therefore, we first transform the point from the full image coordinates to the cropped coordinates, based on the center and scale of the bounding box. The cropped image coordinates are then resized to a fixed resolution and normalize between -1 to 1.  Finally, the 2D keypoint loss, $\mathcal{L}_{j2d}$, is computed based on the normalized 2D keypoints, $\mathbf{J}_{2d}^{\text{norm}}$.

\begin{equation}
\mathcal{L}_{j2d} = \|\hat{\mathbf{J}}_{2d}^{\text{norm}} - \mathbf{J}_{2d}^{\text{norm}}\|_2^2
\end{equation}

\section{CamSMPLify}
Here we provide more details about the optimization procedure used for generating our pseudo ground truth data for 4DHumans dataset. As describe in Eq.~\ref{main:eq:reg} from the main paper, we minimize the energy term $E(\beta, \theta, t^{\text{full}})$ by optimizing for SMPL shape $\beta$ and pose $\theta$ as well as the camera translation $t^{\text{full}}$ in 2 steps.

Initially, we optimize for $n$ iterations focusing solely on the parameters $\beta$ (shape), $\theta_0$ (global orientation), and $t^{\text{full}}$ (translation), while excluding $\theta$ (pose parameters) from the optimization. During this stage, we set the pose prior weight $\lambda_{\text{int}}$ to 0, as we are not yet optimizing the pose. This strategy prevents the model from distorting the body pose excessively to match the keypoints, ensuring that any discrepancies due to incorrect orientation, shape, or translation are addressed first. By refining these parameters initially, we avoid overcompensating for errors related to the pose. After this stage, we update $V_{\text{int}}$ with the output vertices from this optimization stage.
In the subsequent stage, we optimize all the parameters, including $\theta$, and set $\lambda_{\text{int}}$ to 1.0 to obtain our final pose, shape and camera translation. For more details on the hyperparameter settings, please refer to the code.

\section{Shape Evaluation}
Most HPS evaluation benchmarks primarily represent average body shapes and offer limited shape diversity, which restricts their effectiveness in assessing improvements in shape accuracy. To address this, we utilize the SSP-3D~\cite{ssp} dataset, which includes a broad spectrum of body shapes. SSP-3D contains 311 real-world images of 62 individuals in fitted clothing, along with estimated ground-truth body shape data.

We evaluate shape accuracy using the PVE-T-SC metric on the SSP-3D dataset. PVE-T-SC, or Per-Vertex Error in T-pose after Scale Correction, calculates the per vertex average error by comparing a reconstructed 3D body mesh in a standardized T-pose to the ground truth. Before computing this error, the scale of the predicted model is adjusted to match the ground truth, ensuring that the metric reflects inaccuracies in shape and pose, rather than differences in overall scale.

As demonstrated in Table~\ref{ssp-eval}, CameraHMR outperforms all other benchmarks in terms of shape accuracy, even surpassing methods specifically trained to enhance shape prediction. Additionally, incorporating our improved 4DHumans pGT into the training process, along with BEDLAM, further improves shape accuracy, highlighting the high quality of the shape information in our pseudo ground truth.
\begin{table}
\centering
{\scriptsize
  \begin{tabular}{l|c|c}
    \toprule
     Method & Model & PVE-T-SC $\downarrow$\\
    \midrule
    HMR~\cite{Kanazawa2018_hmr} & SMPL & 22.9\\
    SPIN~\cite{spin} & SMPL & 22.2 \\
    SHAPY~\cite{shapy} & SMPL-X & 19.2\\
    STRAPS~\cite{ssp} & SMPL & 15.9\\
    Sengupta et. al~\cite{sengupta} & SMPL & 13.6\\
    CLIFF~\cite{li2022cliff} & SMPL & 18.4\\
    BEDLAM-CLIFF~\cite{bedlam} & SMPL-X & 14.2\\
    CameraHMR (B) & SMPL & 13.3\\
    CameraHMR (B+4DH)& SMPL & \textbf{11.6}\\
    
    \bottomrule
  \end{tabular}
  }
  \caption{Shape error evaluation on SSP-3D dataset. B means trained on BEDLAM and AGORA and 4DH means trained on 4DHumans.}
  
\label{ssp-eval}

\end{table}

\section{More Qualitative Results}
\begin{figure*}
\centerline{\includegraphics[width=1\textwidth]{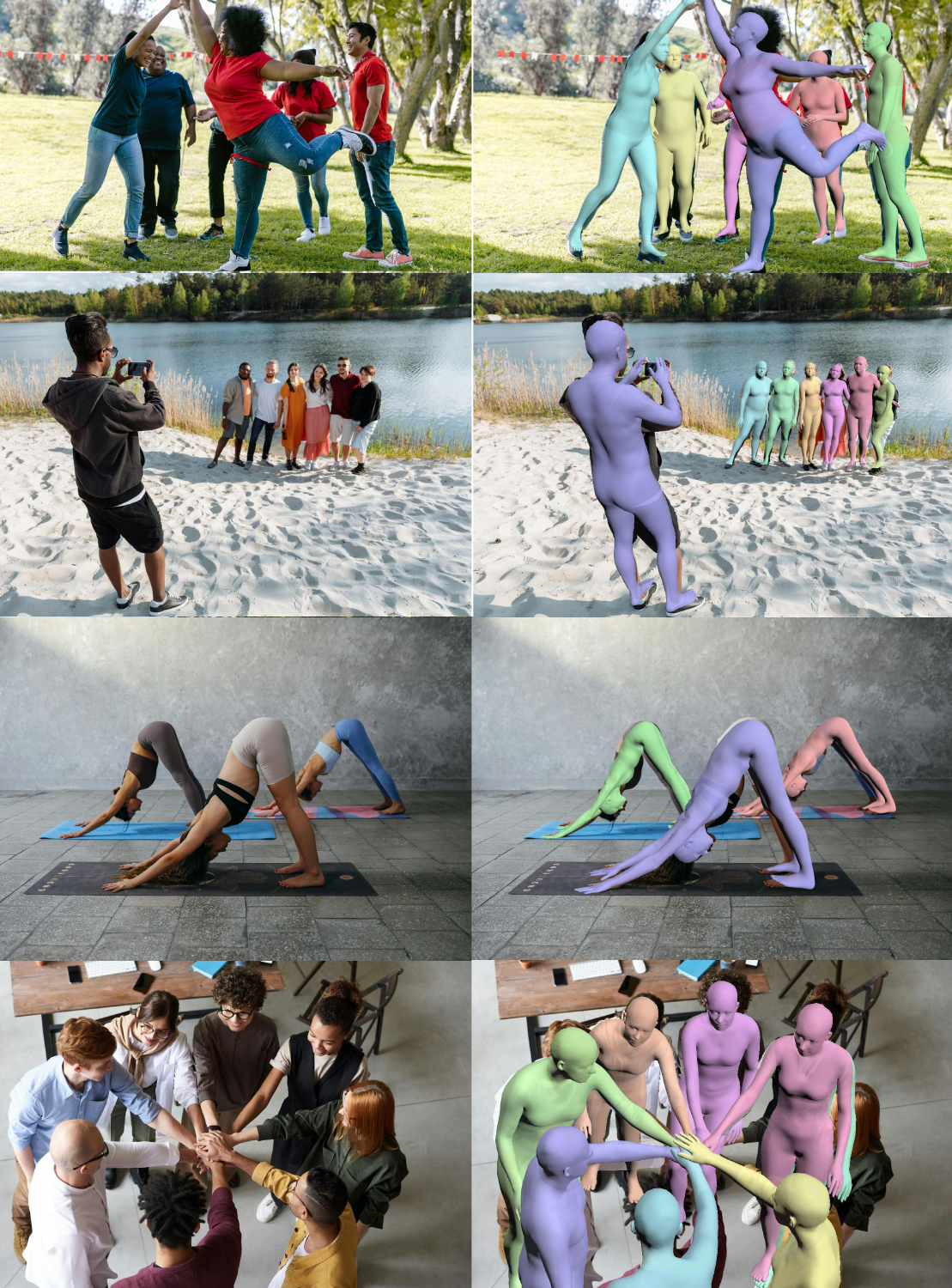}}
\caption{CameraHMR results on landscape images downloaded from Pexels~\cite{pexel}.}
\label{fig:qualitative_landscape}

\end{figure*}

\begin{figure*}
\centerline{\includegraphics[width=0.9\textwidth]{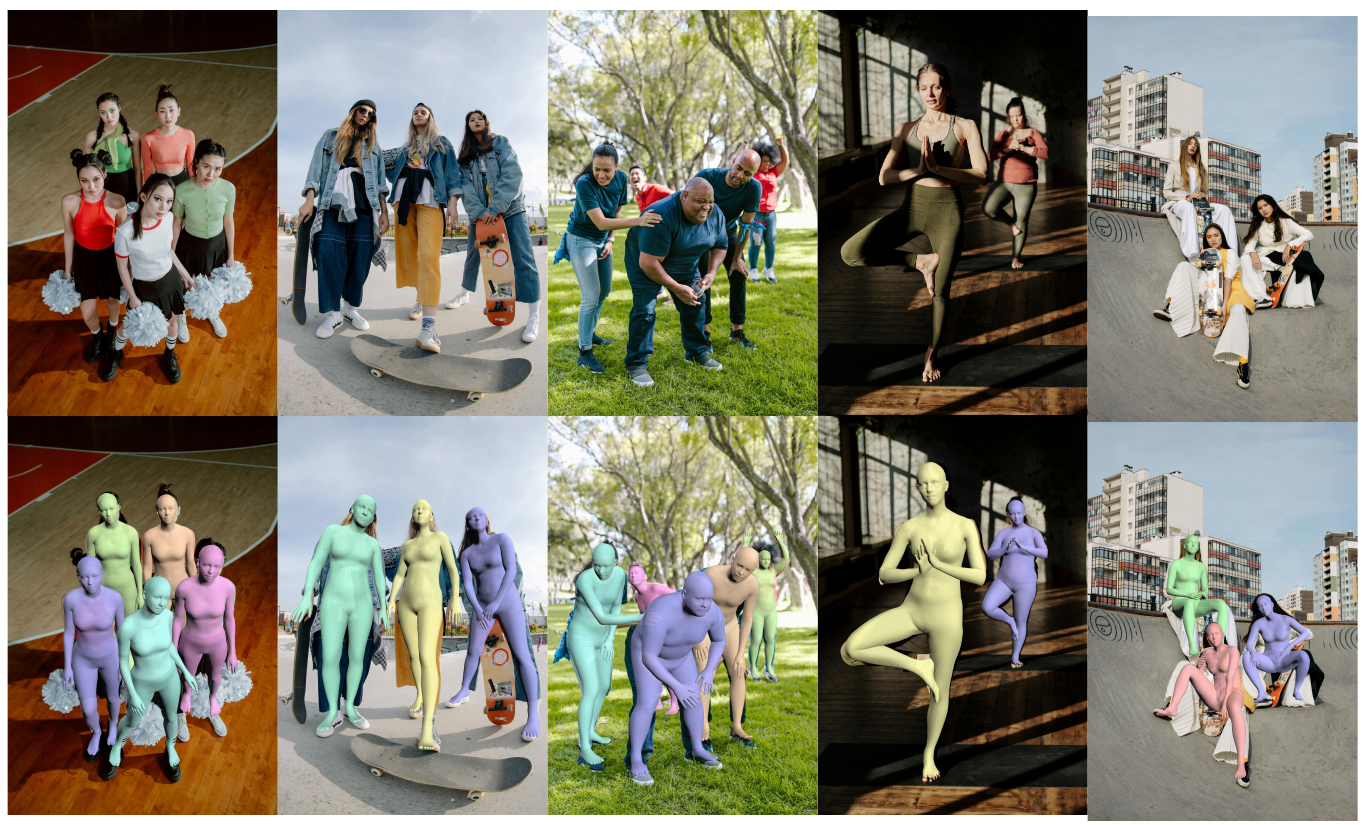}}
\caption{CameraHMR results on portrait images downloaded from Pexels~\cite{pexel}.}
\label{fig:qualitative_portrait}

\end{figure*}

In Fig.~\ref{fig:qualitative_landscape} and Fig.~\ref{fig:qualitative_portrait}, we present qualitative results of CameraHMR applied to images downloaded from Pexels~\cite{pexel}. The results for multi-person images are obtained by first generating the bounding box for each person using Detectron2~\cite{detectron2} on the full image. These cropped bounding boxes are then fed into CameraHMR. The results demonstrate that CameraHMR effectively estimates both accurate body poses and detailed body shapes, even for complex body poses and camera angles.

\begin{figure*}
\centerline{\includegraphics[width=0.9\textwidth]{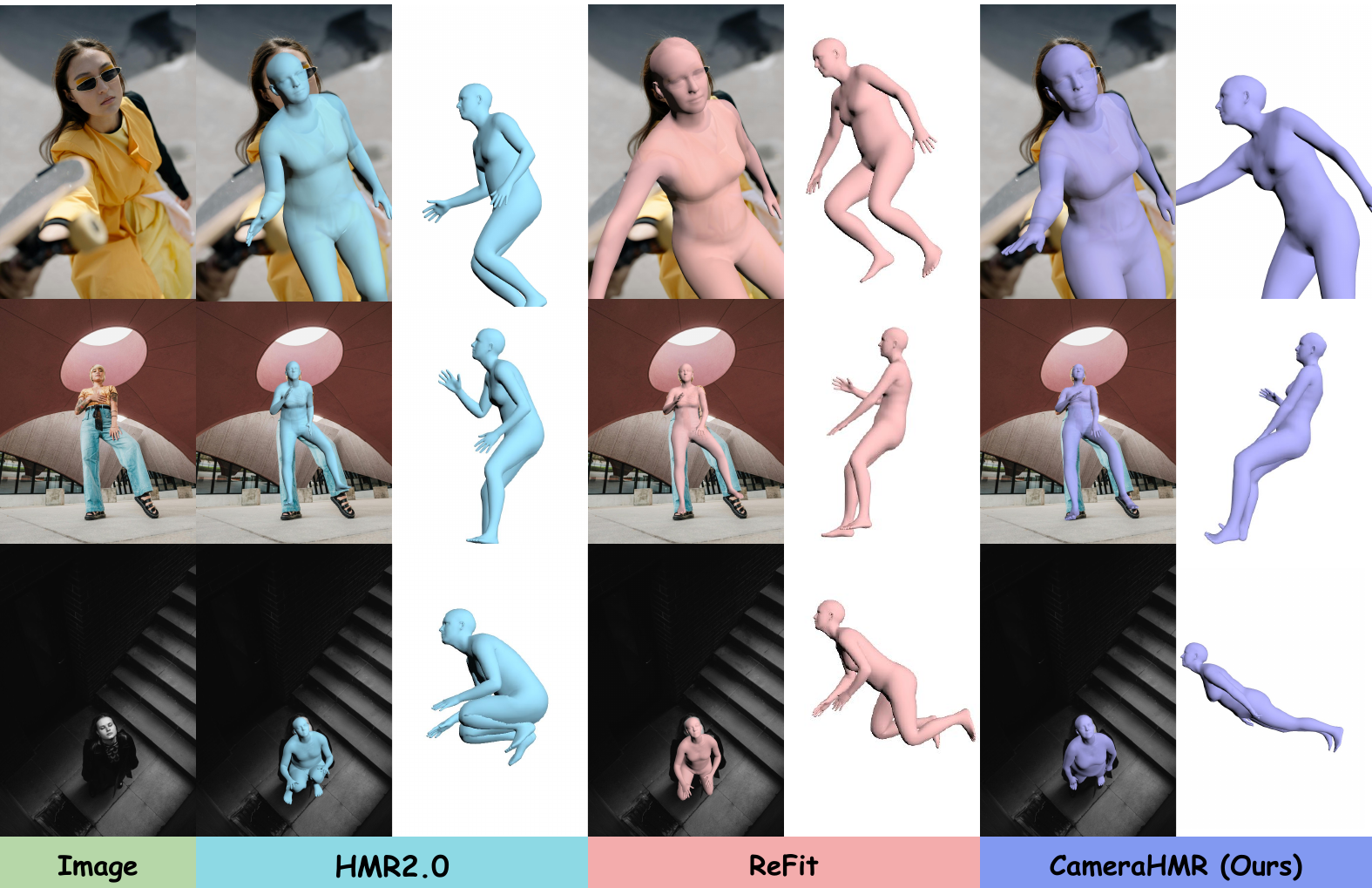}}
\caption{CameraHMR achieves more accurate 3D pose estimation, shape reconstruction, and 2D alignment with the image even for extreme camera angles, outperforming other methods in these challenging scenarios.}
\label{fig:qualitative_comp}
\end{figure*}

In Figure~\ref{fig:qualitative_comp}, we compare the results of CameraHMR with HMR2.0~\cite{hmr2} and ReFit~\cite{refit} on images from Pexels. HMR2.0 employs a weak perspective camera model, while ReFit, like CameraHMR, uses a full perspective camera model during training. Despite achieving good alignment on images with standard focal length, HMR2.0 often results in unrealistic body poses. 2D alignment with the image also gets worse as FoV of the image increases as shown in some of the images in Figure~\ref{fig:qualitative_comp}. ReFit~\cite{refit}, although producing a more accurate 3D pose, suffers from poor 2D alignment due to reliance on default camera parameters during inference. In contrast, CameraHMR leverages robust camera intrinsics predicted by our HumanFoV model, resulting in accurate body poses and shapes as well as improved 2D alignment, even under extreme camera conditions.

\end{document}